\documentclass[runningheads]{llncs}
\usepackage{graphicx} 
\usepackage{amsmath}
\usepackage{makecell}
\usepackage{hyperref}
\usepackage{color}
\usepackage{colortbl}

\newcommand{\alex}{\texttt{AlexNet} }
\newcommand{\google}{\texttt{GoogLeNet} }

\definecolor{grd}{rgb}{0.4, 0.9, 0.35}
\definecolor{grl}{rgb}{0.6, 1, 0.6}
\definecolor{grll}{rgb}{0.8, 1, 0.8}
\definecolor{ye}{rgb}{1, 1, 0.6}
\definecolor{or}{rgb}{1, 0.9, 0.7}
\definecolor{re}{rgb}{1, 0.7, 0.6}

\begin{document}
\title{Evaluating CNNs on the Gestalt Principle of Closure}
%
%
\author{Gregor Ehrensperger \and
Sebastian Stabinger \and
Antonio Rodr\'iguez S\'anchez}
%
%
\institute{University of Innsbruck, Technikerstraße 21a, 6020 Innsbruck, Austria \url{https://iis.uibk.ac.at/}}

\maketitle              
\begin{abstract}
Deep convolutional neural networks (CNNs) are widely known for their outstanding performance in classification and regression tasks over high-dimensional data. This made them a popular and powerful tool for a large variety of applications in industry and academia. Recent publications show that seemingly easy classifaction tasks (for humans) can be very challenging for state of the art CNNs. An attempt to describe how humans perceive visual elements is given by the Gestalt principles. In this paper we evaluate \alex and \google regarding their performance on classifying the correctness of the well known Kanizsa triangles, which heavily rely on the Gestalt principle of closure. Therefore we created various datasets containing valid as well as invalid variants of the Kanizsa triangle. Our findings suggest that perceiving objects by utilizing the principle of closure is very challenging for the applied network architectures but they appear to adapt to the effect of closure.

\end{abstract}

\section{Introduction and Related Work}
Convolutional neural networks have gained enormous interest in industry and research over the past years because they provide outstanding performance in many visual classification tasks. The basic architecture of a CNN was first introduced by LeCun et al. \cite{lecu1989} in 1989. Almost a decade later LeCun et al. \cite{lecu1998} created \texttt{LeNet-5} which was able to classify handwritten digits with an accuracy exceeding 99\% on the MNIST dataset. In 2012 Krizhevsky et al. managed to train a deep CNN -- later known as \alex -- to classify 1.2 million images into 1000 different classes with an impressive top-5 test error rate of 15\%. In 2014 Szegedy et al. \cite{szeg2015} introduced the famous \texttt{Inception} architectures which are also known as \google in hommage to \texttt{LeNet}. In this paper we experiment with \alex and \texttt{GoogLeNet}\footnote{In our case: \texttt{Inception~v3}}.

\textit{Gestalt psychology} explains different perceptual phenomena. In 1923 Wertheimer \cite{wert1923} described a set of rules which are essential for our perception of objects, the so-called \textit{Gestalt principles}. One of these principles is given by the \textit{principle of closure}, which states that humans tend to fill visual gaps to perceive objects as being whole, even when fragments are missing. Another principle is given by the \textit{principle of similarity} which states that shapes that are similar to each other tend to be perceived as a unit. Stabinger et al. \cite{stab2016} could show that neither \texttt{LeNet} nor \google are capable of comparing shapes. Further experiments were performed in the context of the \textit{principle of symmetry}, where Stabinger et al. \cite{stab2017} found variations of a dataset which seem to be at the border of what CNNs can do.  Kim et al. \cite{kim2019} just recently adapted tools which are used in psychology to study human brains to analyse the neural responses within CNNs to see whether they utilize the \textit{principle of closure}. They showed that under certain circumstances neural networks do respond to closure effects. In this paper we also experiment with the \textit{principle of closure} by evaluating the ability of the given CNNs to decide whether an image contains a valid or an invalid Kanizsa triangle. Our goal is to gain more insight from a practical point of view into how challenging it is for a CNN to exploit the closure effect and how well it performs.
\section{Evaluating CNNs on Datasets Utilizing the Principle of Closure}
\subsection{Kanizsa Triangle Dataset}
\begin{figure}
	\centering
	\includegraphics[width=0.9\textwidth]{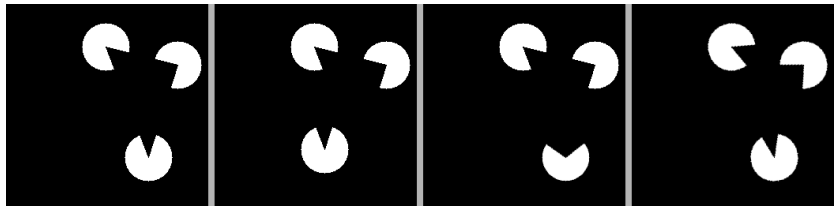}
	\caption{Illustration of the different datasets. From left to right: Kanizsa triangle without error, \textbf{OFFS}, \textbf{ANGLE}, \textbf{ROT}.}
	\label{img:coll}
\end{figure}

For our tests we generated datasets with 50.000 images each\footnote{30.000 training, 10.000 validation and 10.000 test images; dimension: $256 \times 256$ px}. One half of each dataset shows valid Kanizsa triangles, the other half invalid variants of the Kanizsa triangle. We created the following scenarios (see \hyperref[img:coll]{Fig.~\ref{img:coll}}).
\begin{enumerate}
	\item \textbf{OFFS}: Translate one of the vertices by a random offset\footnote{To maximize the visual error, the offset is applied in the direction of the connecting line of the other two vertices.}.
	\item \textbf{ANGLE}: Change the opening angle within one vertex.
	\item \textbf{ROT}: Rotation of one to three vertices by a random angle.
	\item \textbf{COMB}: Each invalid Kanizsa triangle contains exactly one of the errors out of the set $\{\operatorname{\bf{ROT}}, \operatorname{\bf{ANGLE}}, \operatorname{\bf{OFFS}}\}$.
\end{enumerate}
\subsection{Evaluating the CNNs}
	It took \alex 140 and \google 99 epochs\footnote{All CNNs were trained using NVIDIA DIGITS \url{https://developer.nvidia.com/digits} with the Torch backend and  def. settings: fixed learning rate = 0.01, solver = SGD.} to correctly classify 95\% of the validation set of \textbf{COMB}, which suggests that this problem is rather challenging for the observed CNN architectures. As a reference we used a subset\footnote{While the original MNIST dataset contains 60.000 training images and 10.000 validation images, we moved 10.000 training images to a test image set, and deleted 20.000 of the training images while not changing the distribution of the images among the classes. We did this to be comparable with our own datasets.} of the MNIST dataset to train \alex and \texttt{GoogLeNet}.  It took \alex only one epoch and \google twelve epochs to achieve a test error rate of less than 5\%. Please note that one needs to distinguish between ten different classes in the MNIST dataset, while the datasets in this paper only consist of two classes. 
	
To gain more insight, we split the problem into its components and look at the number of epochs the CNNs need to accomplish an accuracy of at least 95\% on the validation set\footnote{Trained multiple times; lowest results are displayed.}:
\begin{enumerate}
	\item \textbf{OFFS}: \alex 27 epochs, \google 188 epochs.
	\item \textbf{ANGLE}: \alex 12 epochs, \google 6 epochs.
	\item \textbf{ROT}: \alex 6 epochs, \google 5 epochs.\footnote{In further experiments we also worked with a set where only one of the three vertices was rotated. \alex needed at least 109 epochs to be able to classify 95\% correctly, while \google needed 6 epochs.}
\end{enumerate}
We observe that \google needs many epochs to reach sufficient accuracy on \textbf{OFFS}, and \alex requires significantly more epochs to learn to classify \textbf{OFFS} than to correctly classify \textbf{ANGLE} and \textbf{ROT}. Furthermore, we want to point out that \alex needed many trials before being able to find a satisfying classifier on the problems involving angles, while \google needed many trials on the \textbf{OFFS} dataset.
\section{Interpretation and Further Results}
Although the problems seem to be very similar at first glance, \textbf{ANGLE} and \textbf{ROT} are locally solvable. Basically it suffices to detect the opening angle and its orientation for each vertex. Then, without considering the position of the vertices, comparing these features leads to the classification result. For \textbf{OFFS} the CNN needs to use higher-level features since it is not able to decide locally anymore. To make this more evident, if we consider the classification matrices in \hyperref[tab:summ]{Table~\ref{tab:summ}}, we observe that:
\begin{table}[htbp]
    \centering
\caption{Summary of test error rates on various datasets [\%]. We trained the CNNs until they converged and evaluated the test sets with a model where the losses on the train and validation set became more stable. The number of epochs that were required to reach this state are indicated in the second column. }
    \label{tab:summ}
\begin{tabular}{|c|c|c|ccc|}
\hline
& \# epochs & \diaghead{MMMMMMMMM}{trained on}{tested on} & \textbf{OFFS} & \textbf{ANGLE} & \textbf{ROT} \\
    \hline
\alex   & 80 & \textbf{OFFS}   & \cellcolor{grd}0.7  & \cellcolor{grl}6.4 & \cellcolor{grl}10.1 \\
\alex   & 20 & \textbf{ANGLE}  & \cellcolor{re}48.8 & \cellcolor{grd}1.2  & \cellcolor{grd}4.8  \\
\alex   & 8 & \textbf{ROT} & \cellcolor{re}50 & \cellcolor{grd}4.3 & \cellcolor{grd}0.3  \\
\hline
\hline
\google & 204 & \textbf{OFFS}   & \cellcolor{grd}0.7  & \cellcolor{grd}1.4 & \cellcolor{grl}14.5   \\
\google & 8 & \textbf{ANGLE}  & \cellcolor{re}50 & \cellcolor{grd}0  & \cellcolor{grd}0  \\
\google & 6 & \textbf{ROT} & \cellcolor{re}50 & \cellcolor{grd}1  & \cellcolor{grd}0.1  \\
\hline
\end{tabular}%
\end{table}        
\begin{enumerate}
\item CNNs trained on \textbf{ANGLE} are also able to classify \textbf{ROT} and vice versa, but they are not able to classify \textbf{OFFS} above chance.
\item CNNs trained on \textbf{OFFS} are able to classify \textbf{ANGLE}, as well as \textbf{ROT} above chance although they did not encounter any of these problems before.
\end{enumerate}
\section{Conclusion}
Our findings suggest that in order to discriminate classes in which the positions of the objects matter, the CNNs need to detect higher-level features and generalize. Apparently, CNNs trained on such a problem set also have significantly lower test error rates on previously unseen perturbations of the data, which suggests that -- in our case -- they are exploiting the principle of closure. Our experiments show that training on these features is quite challenging, needing significantly more epochs than training on the MNIST dataset. We believe that further analysis of the performance of CNNs in the context of the Gestalt principles is a promising area for future research to gain a better understanding of the differences and similarities between human and artificial neural network perception.

\end{document}